\documentclass[letterpaper, 10 pt, conference]{ieeeconf}  

\IEEEoverridecommandlockouts                              

\usepackage{booktabs}
\usepackage{siunitx}
\usepackage{wrapfig}
\usepackage{tabularx}
\usepackage{times}
\usepackage{xr}
\makeatletter

\usepackage{multicol}
\usepackage{amssymb}
\usepackage{amsmath}
\usepackage{bm}      
\usepackage{subfig}
\usepackage{graphicx}
\usepackage{caption}
\usepackage{floatrow}
\usepackage{flushend}
\usepackage{cite}
\usepackage{algorithm}
\usepackage{algpseudocode}
\usepackage{titlesec}

\DeclareMathOperator*{\argmax}{arg\,max}

\algdef{S}[FOR]{ForEach}[1]{\algorithmicforeach\ #1\ \algorithmicdo}
\algnewcommand\algorithmicinput{\textbf{Input:}}
\algnewcommand\INPUT{\item[\algorithmicinput]}
\algnewcommand\algorithmicoutput{\textbf{Output:}}
\algnewcommand\OUTPUT{\item[\algorithmicoutput]}

\usepackage{xcolor}

\title{
Causal Inference for De-biasing Motion Estimation from Robotic Observational Data
}

\author{Junhong Xu, Kai Yin, Jason M. Gregory, Lantao Liu
\thanks{J. Xu and L. Liu are with the Luddy School of Informatics, Computing, and Engineering  at Indiana University, Bloomington, IN 47408, USA. E-mail:
        {\tt\small \{xu14, lantao\}@iu.edu}.
K. Yin is with Expedia Group. E-mail:
        {\tt\small kyin@expediagroup.com}.  
J. Gregory is with U.S. Army Research Laboratory, Adelphi, MD. E-mail: {\tt \small jason.m.gregory1.civ@mail.mil}
}
}

\begin{document}

\maketitle
\thispagestyle{empty}
\pagestyle{empty}

\begin{abstract} 
Robot data collected in complex real-world scenarios are often biased due to safety concerns, human preferences, and mission or platform constraints. Consequently, robot learning from such observational data poses great challenges for accurate parameter estimation. We propose a principled causal inference framework for robots to learn the parameters of a stochastic motion model using observational data. 
Specifically, we leverage the de-biasing functionality 
of the potential-outcome causal inference framework, the Inverse Propensity Weighting (IPW), and the Doubly Robust (DR) methods, to obtain a better parameter estimation of the robot's stochastic motion model. 
The IPW is a re-weighting approach to ensure unbiased estimation, and the DR approach further combines any two estimators to strengthen the unbiased result even if one of these estimators is biased.
We then develop an approximate policy iteration algorithm using the bias-eliminated estimated state transition function.
We validate our framework using both simulation and real-world experiments, and the results have revealed that the proposed causal inference-based  navigation and control framework can correctly and efficiently learn the parameters from biased observational data. 
\end{abstract}

\vspace{-5pt}
\section{Introduction}
Modern robots gain complex skills by leveraging existing robotic data. 
However, the data collection processes are often biased due to robotic safety issues, human preferences, or system constraints. This results in a gap between many data-driven approaches and the target robotic applications.
Take the mobile robot navigation as an example, we generally disallow the robot to randomly explore the environments
especially in complex, cluttered, or unstructured outdoor scenarios.  
The data collected by other agents (human operators or a carefully designed system) that ensure the robot's safety is termed as {\em observational} data.
In this context, the data lack sufficient ``randomness" because the robot motion is directly affected by, and thus biased from, many extraneous factors, which can lead to highly inaccurate parameter estimation and learning results. 
For instance, to train a mobile robot to navigate and control, the observational trajectory datasets can be collected from human-piloted trials/demonstrations, or from unmanned autopilot missions where the robot follows some motion planners exclusive to specific missions. 
In both cases, the data can be highly biased from different humans or missions. 
This is because while operating a robot, humans typically have preferences (e.g., due to safety concerns) over control strategies under different environmental conditions, and the robot motion trajectories can also vary significantly even for the same control strategy but with differing mission constraints (e.g., with vs. without a map as prior knowledge). 
The bias caused by different platforms is also important.
A large vehicle may be able to ignore small bumps whereas a small vehicle might choose to avoid  them. 
If we use (or leverage) the data collected by large vehicles to train small robots, the models and behaviors may not be transferrable. 
Additionally, the data collection processes are typically unknown, and one cannot always infer the processes as data was collected in the past. 


In many scenarios, the bias can be hardly eliminated while collecting data. 
Our objective is the ``de-biased" learning from the biased observational data.
We propose to design a fundamental causal inference framework for autonomous systems to learn parameters of stochastic motion using offline observational data.
Since the decision-making of a robot moving in unstructured environments typically requires the robot to account for uncertain action (motion) outcomes and meanwhile maximize the long-term return, we base our formulation on the Markov Decision Process (MDP) which has been shown as a powerful framework for formulating robot decision making problems~\cite{boutilier1999decision}. 

Our work implicitly builds upon the Neyman-Rubin causal model~\cite{imbens2015causal, guo2020survey}, 
and we integrate the {\em causal effect} into a continuous-state MDP. 
The resulting state of action can be viewed as the potential outcome. 
In this work, we leverage a diffusion approximation to MDP for stochastic motion control~\cite{xu2020kernel}, which allows us to narrow our attention to the estimation of only the first and second moments of robot stochastic state transition for every action rather than estimating the complete and exact form of the distribution.
This reduces the complexity of the original problems and allows us to draw on existing causal inference approaches most of which also concentrate on exploiting the first two moments of the potential outcomes in the sample space. 
Our contributions are summarized as follows:
\begin{itemize}
[leftmargin=20pt] \vspace*{-2pt}
\setlength\itemsep{-1pt}
    \item First, to reduce the bias in offline observational data, we apply the Inverse Propensity Weighting (IPW) method to estimate the first and second moments of transition probabilities.
    Different from existing work where typically binary actions are assumed/used, we generalize the methods to multiple 
    actions in the robotics context. 
    \item Second, given any regression model, we improve the estimation by combining together the IPW and regression estimators. Such an approach possesses the Doubly Robust (DR) property in the sense that if either the propensity score model or the regression is incorrect, the final estimation still remains unbiased (in subpopulation) as long as the other model is correct. 
    \item Finally, we develop an efficient policy iteration algorithm 
    that can seamlessly integrate IPW or DR methods in the diffusion approximate MDPs which only require the first and second moments of the state transition functions. The algorithm is able to correctly and efficiently learn the parameters in the diffusion approximate MDPs from biased observational data. 
\end{itemize}
    Extensive simulated experiments and real-world experiments on rough terrains show that the policy iteration algorithm equipped with causal inference-based model-learning generates safer navigational behaviors than the baseline methods.

\vspace{-6pt}
\section{Related Work and Background}

This paper focuses on learning motion models 
using observational or offline data. A major problem involved in observational or offline data  is that the collection mechanisms are typically unknown to the users. 
This may cause potential bias if researchers directly estimate a model from these data. 
The vast literature on causal inference has already recognized the difficulties in identifying and estimating action (treatment) effects and researchers have developed tools for estimating these effects in different research communities \cite{rubin1974estimating, yao2020survey}. 
Yet using causal inference to solve robotics problems is a new topic and related work is 
scarce. 

\textbf{Neyman-Rubin Causal Model and Structural Causal Model:}
A common causal model in statistics and economics, the so-called Neyman-Rubin causal model, is based on the idea of potential outcomes \cite{rubin2005causal, johansson2020generalization}. 
It assumes that a potential outcome associated with an individual when an action is taken. For example, in a study of a medicine's effect on patients with hypertension, the potential outcomes for this patient are the blood pressure {\em with} and {\em without} the medicine treatment (action), respectively. In practice, the patients are sorted into treatment and control groups for clinical trial reasons, and data scientists may not be able to control the treatment assignment. 
We will be able to observe only one potential outcome, the {\em factual} outcome (e.g., after taking real medicine); we cannot obtain 
the {\em counterfactual} potential outcome for that patient at the same time (e.g., taking a placebo instead of real medicine).
Here the ``counterfactual"  means the result in the other scenario is often not allowed or possible to obtain. 
The problem is, to measure the causal effect, we need to compare {\em both the factual and counterfactual} potential outcomes for the {\em same patient at the same time}. 
The Neyman-Rubin causal model concerns the treatment assignment processes and aims to estimate the population-level average causal effect by inferring the counterfactual potential outcomes. Obviously, certain assumptions about treatment assignment need to be imposed to identify the causal effect \cite{angrist1995identification}. 
Under these assumptions, various methods based on the propensity score, matching methods, or tree-based models have been investigated \cite{rosenbaum1983central, athey2015machine, wager2018estimation}.
In this work, we leverage the de-biasing functionality of the potential-outcome framework to obtain better estimations of the robot motion model outcome (which can be viewed as a causal effect of the robot's action) from observational data. 
Other de-biasing methods have also been proposed in the literature~\cite{sagawa2020investigation}, but these methods do not consider action selection strategy or continuous features.
Another popular framework in artificial intelligence or machine learning communities is the causal structural models based on directed acyclic graphs \cite{pearl2009causality,colombo2012learning, kocaoglu2017experimental}. This framework describes causal relations by graphs and employs a set of simultaneous structural equations to detect the causal effect \cite{peters2017elements, pearl2009causal}. The above two causal frameworks are complementary and appropriate for different questions \cite{imbens2020potential}. The relevant methods are employed in reinforcement learning~\cite{jiang2016doubly, kallus2020deepmatch, dudik2011doubly}, recommendation systems~\cite{schnabel2016recommendations}, and computer vision~\cite{lopez2017discovering}.

\textbf{Offline Reinforcement Learning:} Our work is complementary to the recently rising offline reinforcement learning (RL)~\cite{levine2020offline, rakelly2019efficient, kumar2020conservative, moerland2020model}, in particular, offline model-based  RL~\cite{yu2020mopo, moerland2020model, kidambi2020morel} that leverages previously collected data or available logs to learn a transition model for planning.
We focus on leveraging the offline data because the conventional online (deep) model-free and model-based reinforcement learning (RL) methods~\cite{sutton2018reinforcement, kober2013reinforcement} attempt to explore the environment and utilize data collected online to improve the model, the value function, or the policy, which incurs risk in field robotics as data collection is expensive and with safety risks.
On the other hand, learning from offline data ensures safety during robot learning because there is no online interaction with the environment. 
Most constraint-based offline model-based RL methods learn a transition model using the standard (regression) techniques (without de-biasing) and penalize the unknown (infrequently visited) state-action space~\cite{yu2020mopo, moerland2020model, kidambi2020morel} during planning, while our work focuses on causal inference methods to de-bias the transition model estimates.
Thus, causal-based methods can be used to improve the constraint-based offline model-based RL by improving the model estimation in the observed part of the state-action space from the data, which we demonstrate in Secion~\ref{sec:simulated-exp}.

\textbf{Contextual Markov Decision Processes:}
We model our decision-making problem by Contextual Markov Decision Processes (MDPs) $\{\mathcal{S}, \mathcal{A}, P, R, \mathcal{C}\}$~\cite{hallak2015contextual}, where the first four elements are state space, action space, transition probability measure, and reward function, respectively~\cite{hallak2015contextual, theocharous2001learning}.
The additional element $\mathcal{C}$ denotes a set of environmental feature vectors. 
In mobile robot navigation, for example, a state can include the robot's pose and body velocity, and the features can be associated with terrain type and roughness (grass, sand, rocks), terrain elevation (hills, cliffs), and wheel traction with respect to the surface roughness.
These environmental features affect the transition probability but cannot be controlled by robots.
Accordingly, the function of transitioning to the next state $s'$ from $s$ is dependent on features $c\in \mathcal{C}$ and is written as $p(s'|s, a; c)$, where $p$ is the density (or mass) function of measure $P$.

We consider the class of deterministic policies $\pi \in \Pi: \mathcal{S}\times\mathcal{C} \rightarrow \mathcal{A}$, which is a mapping from state and feature to action. To simplify the notation, we will write $\pi(s)$ instead of $\pi(s, c)$. And the same convention is applied to the value function and reward function. 
We consider the infinite-horizon case, and the value function at any state  $s$ is defined as $v^{\pi}(s) = \mathbb{E}^{\pi}[\sum_{k=0}^{\infty} \gamma^k R(s_k, \pi(s_k)) | s_0 = s]$,
where $\gamma$ is the discount factor. 
The above equation can be written recursively as 
\begin{equation}\label{Bellman-Eqn}
    v^{\pi}(s) = R(s, \pi(s)) + \gamma\,\mathbb{E}_{s'\sim p}^{\pi}[v^{\pi}(s')\mid s].
\end{equation}
The goal of the robot is to find an optimal policy that maximizes the value function at every state $\pi^{*}(s) = \argmax_{\pi \in \Pi} \big \{ R(s, \pi(s)) + \gamma\, \mathbb{E}_{s'\sim p}^{\pi}[v^{\pi}(s')|s]\big\}$.
The above definitions clearly show $v(\cdot)$, $R(\cdot, \cdot)$, and $\pi(\cdot)$ depend on the feature because the transition probability function has a dependence on the feature.
In many real-world scenarios, the transition function $p(s'|s, a; c)$ is unknown before deploying the robots to the field, making it impossible to obtain the optimal policy.
In this work, we demonstrate how to \textit{correctly} utilize the offline observational data to estimate the transition and then compute the optimal policy.

\section{Methodology}
We first formulate a diffusion approximated MDP framework~\cite{xu2020kernel} where the main learning task is to estimate the first and second moments of transition functions (Sect.~\ref{DiffApproxMDP}).
We then develop two principled learning methods for the offline data by causal inference approaches (Sect.~\ref{sec:IPW} and~\ref{sec:DR}). 
\vspace{-5pt}
\subsection{Diffusion Approximation to {Contextual} MDP} \label{DiffApproxMDP}
It is difficult to learn an exact state transition probability distribution for robotic systems in complex unstructured environments. 
Thus, we opt to build upon a diffusion-approximated MDP which computes an approximation to the optimal value function only using the first and second-order moments of the transition probability. 

We consider a continuous $k$-dimensional state space $\mathcal{S}$ and a finite set of actions $\mathcal{A}$.  
Suppose that the value function $v^{\pi}(s)$ for any given policy $\pi$ has continuous first and second order derivatives. 
We subtract both hand-sides by $v^{\pi}(s)$ from Eq.~\eqref{Bellman-Eqn} and 
then take Taylor expansions of value function around $s$ up to second order: 
\begin{align}
\resizebox{0.45\textwidth}{!}{${\displaystyle
    \gamma\,\Big((\mu^{\pi}_s)^T\,\nabla v^{\pi}(s) +\frac{1}{2}\nabla\cdot \sigma_s^{\pi}\nabla v^{\pi}(s)\Big) - 
    (1-\gamma)\,v^{\pi}(s) \simeq -R(s, \pi(s))}$}
    \label{bellman-type-pde}
\end{align}
where $\nabla$ is vector differential operator; $\mu^{\pi}_s$ and $\sigma^{\pi}_s$ are the first moment (a $k$-dimensional vector) and the second moment (a $k$-by-$k$ matrix) of transition functions, respectively, with the following form $ (\mu^{\pi}_s)_i = \int p(s'| s,\pi(s); {c})(\Delta s)_i\,ds', \,(\sigma^{\pi}_s)_{i, j} = \int p(s' | s,\pi(s); {c})(\Delta s)_i(\Delta s)_j\,ds',$
where $(\Delta s)_i$ denotes the $i$-th component of $s' - s$.
Because it is generally impossible to represent the value function for an infinite number of states over the continuous state space, we represent the value function at any state $s'$ by its values at only a predefined finite number of {\em supporting states} $\mathbf{s}=\{s^{1}, \ldots, s^N\}$. Such representation is done through a kernel approximation $v^{\pi}(s') = \mathbf{k}(s', \mathbf{s})^T\,\left(\lambda\mathbf{I}+\mathbf{K}\right)^{-1}\,V^{\pi}$,
where $k(\cdot, \cdot)$ is a generic kernel function~\cite{hofmann2008kernel};
$\lambda \geq 0$ is a regularization factor;
$\mathbf{K}$ is the Gram matrix with $(i, j)$-th entry 
$k(s^i, s^j)$;
$\mathbf{k}(\cdot, \mathbf{s})$ is a column vector with $i$-th component $k(\cdot, s^i)$;
and 
$V^{\pi}$, a $N\times 1$ vector with $i$-th component $v^{\pi}(s^i)$,
is state-values at the supporting states and needs to be computed.
The following linear system is derived to compute $V^{\pi}$: 
$\left(\mathbf{M}^{\pi} \left(\lambda\mathbf{I}+\mathbf{K}\right)^{-1} -(1-\gamma)\,\mathbf{I}\right)\,V^{\pi} = \mathbf{R}^{\pi},$
where $\mathbf{I}$ is an identity matrix, $\mathbf{R}^{\pi}$ is a 
vector with $i$-th element $-R(s^i, \pi(s^i))$, and $\mathbf{M}^{\pi}$ is a matrix whose $(i, j)$-th entry is $\gamma((\mu^{\pi}_{s^i})^T\nabla_{s^i} + \frac{1}{2}\nabla_{s^i}\cdot \sigma_{s^i}^{\pi}\nabla_{s^i})k(s^i, s^j).$

When the two moments are {\em known}, the optimal policy solution can be computed by the {\em policy iteration} algorithm 
~\cite{XuYinLiu2020kernel,boutilier1999decision}.
Suppose that the value function under the current policy $\pi_t$ is obtained at the $t$ iteration.
We then improve the policy by the following equation to get $\pi_{t+1}$ at every state $s$, 
$\argmax_{a\in\mathcal{A}}\Big\{R(s, a)+\gamma\Big((\mu_s^{a})^T\,\nabla+ 
     \frac{1}{2}\nabla\cdot \sigma_s^a\nabla\Big) v^{\pi_t}(s) \Big\} $.
Since $\mu_s$ and $\sigma_s$ are unknown, next we need to estimate these two parameters 
for each action $a$ from the observational trajectory dataset $\mathcal{D}$.
A similar kernel-based RL method has been proposed in~\cite{taylor2009kernelized}, which uses kernel-based methods (e.g., Gaussian Processes) to approximate the reward function and the transition function. 
In contrast, we leverage the diffusion approximated MDP~\cite{xu2020kernel}, which can be seen as a Laplace approximation to the transition probability distribution.
Then, we use the kernel function to approximate the value function. 
Note that estimation error (bias) due to learning from a biased observational dataset will present regardless of the model choice.
In the following sections, we present two approaches for de-biasing the model estimation.

\vspace{-5pt}
\subsection{Propensity Score Based Causal Inference Approach}\label{sec:IPW}
When estimating the moment functions for action $a$ given the state $s$ and feature $c$ in the dataset $\mathcal{D}$, it is likely or even typical that there are not enough observations with the designated $(s, c)$ for action $a$. Thus, we have to leverage a subset of data with the state and feature {\em close to} $(s,c)$ and also associated with the action $a$. 
If each data in this subset does \textit{not} have the same chance of having action $a$,
directly estimating the first and second moments of the transition function from $\mathcal{D}$ produces undesirable and biased results. 
To de-bias the estimation, we can place less weight on the samples that lead to biased results. To choose such weight, we must take into account how the observations with action $a$ distribute across states and features. That is, we have to
infer the data collection processes. 

We define {\em propensity score} to be the probability of assigning an action $a$ given state $s$ and feature $c$ in the dataset~\cite{imbens2015causal}, $e_{a}(u) = \mathbb{P}[a | u := (s, c)]$,
where $u$ denotes $(s, c)$. 
The propensity score measures the probability of generating a specific action from some specific state and feature in the observational data. 
An implicit assumption here is that the observed state and feature ``govern" the action generation processes. 
For the data collection process with one robot in a particular mission, propensity score might be viewed as a stochastic policy in terms of MDP. 
However, its meaning can be beyond this scope. When the observational dataset includes data logged from multiple robots in different missions, it reflects the overall random action assignment by pulling diverse data together. 

We start estimating the average {\em first} moment of transition functions for action $a$ within a subset around $u=(s, c)$. 
Suppose that a subset $\mathcal{N}(u)$ of the observational data $\mathcal{D}$ is chosen for this task, and this subset can be defined as neighboring samples measured by the Euclidean distance from $u$. We denote the average first moment by $\mu_a(u):=\mathbb{E}_{\text{sb}}^a(\Delta s)$, where the subscript ``sb" denotes the expectation taken in the subpopulation $\mathcal{N}(u)$.
If $e_a(u)$ is known, it can be utilized to weight the data to reduce the bias of estimation. The corresponding estimator, known as Inverse-Propensity Weighting (IPW), reads
\begin{align}\label{IPW-first-moment}
\resizebox{0.6\hsize}{!}{${\displaystyle
    \hat{\mu}^{IPW}_a(u) = \frac{1}{|\mathcal{N}(u)|}\sum_{i\in\mathcal{N}(u)}\mathbb{I}_{a}(a_{i})\frac{\Delta s_{i}}{e_{a}(u_i)},}$}
\end{align}
where 
$|\mathcal{N}(u)|$ is the number of samples in $\mathcal{N}(u)$; $\mathbb{I}_{a}(\cdot)$ is an indicator function for $a$, i.e., $\mathbb{I}_{a}(a_i)=1$ if and only if $a_i=a$; 
$\Delta s_{i} = s_{i}' - s_{i}$ is the observed {\em state shift} of sample $i$ in the state values between the current state $s_i$ and next state $s_i'$ in the data. 
(Depending on the applications, the state shift can be formulated differently. 
For example, for a ground vehicle, the state can be its pose, so the $\Delta s$ represents the amount of translational and rotational displacement.)

In Eq.~\eqref{IPW-first-moment}, the estimator employs the {\em inverse propensity score} as the weight for observed state shifts. Intuitively, if there are many samples generated toward some value $u$ for action $a$, i.e., the corresponding propensity score is large, then the estimator puts small weight when counting these samples into the final estimation. If in a special case, actions are completely randomly assigned regardless of $u$, then all $e_a(u)$ are equal and no weights are actually needed. 
Similarly, the average {\em second} moment estimator can be expressed as
$\hat{\sigma}^{IPW}_a(u) = 
|\mathcal{N}(u)|^{-1}
\sum_{i\in\mathcal{N}(u)}\mathbb{I}_{a}({a_{i}})\frac{\Delta s_{i}\Delta s_{i}^T}{e_{a}(u)}.$
Now we discuss the estimation of $e_a(u)$.
For propensity score $e_{a}(u)$, we should have $\sum_a e_{a}(u) = 1$. 
We use a non-parametric approach to estimate the propensity score as $ \hat{e}_{a}(u) = \frac{p(a)\,p(u|a)}{\sum_{a}p(a)\,p(u|a)},$
where $p(a) = 
|\mathcal{N}(u)|^{-1}\sum_{i\in\mathcal{N}(u)}\mathbb{I}_{a}(a_i)$; $p(u|a)$ is given by the kernel density estimation (KDE) $p(u|a) \propto \sum_{i\in \mathcal{N}(u)}\mathbb{I}_a(a_i)k(c-c_i; h)$,  
where $k$ is any kernel function and $h$ is the lengthscale parameter of the kernel~\cite{shawe2004kernel} 
to get the corresponding estimates.

\vspace{-7pt}
\subsection{Doubly Robust Estimator}\label{sec:DR}
{
\begin{figure}
{
{\label{fig:}\includegraphics[width=0.7\textwidth]{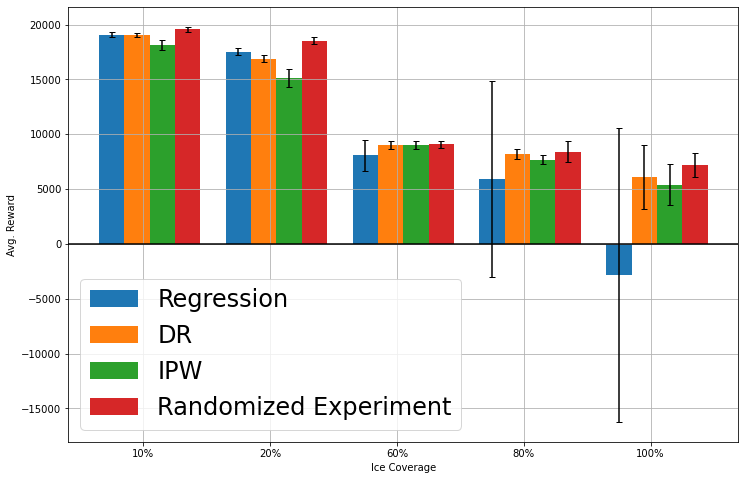}}
} 
{\caption{\small 
The cumulative reward of the four methods averaged over $50$ runs.
    The x-axis and y-axis represent the percentage of ice coverage and the cumulative undiscounted reward, respectively.
    \textit{Regression} denotes the standard regression via KNN; DR and IPW are the proposed causal-based approach. All three methods use observational data.
    The last method, Randomized Experiment, is the behavior policy that uses the data collected by unbiased randomized action selection.  \vspace{-10pt}
  } 
  \label{fig:reward}
  } \vspace{-10pt}
\end{figure} 
}

\begin{figure*}[t] \vspace{-5pt}
    \centering
    \subfloat[]{\label{fig:aggressive-action}\includegraphics[width=0.25\linewidth]{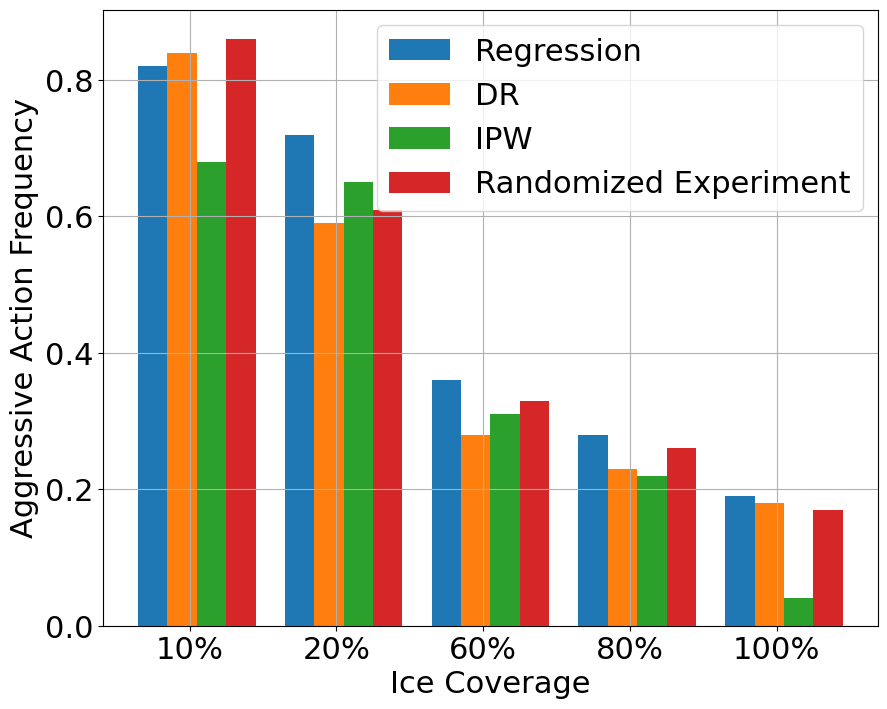}}
    \hspace{0.2in}
    \subfloat[]{\label{fig:linear-v}\includegraphics[width=0.25\linewidth]{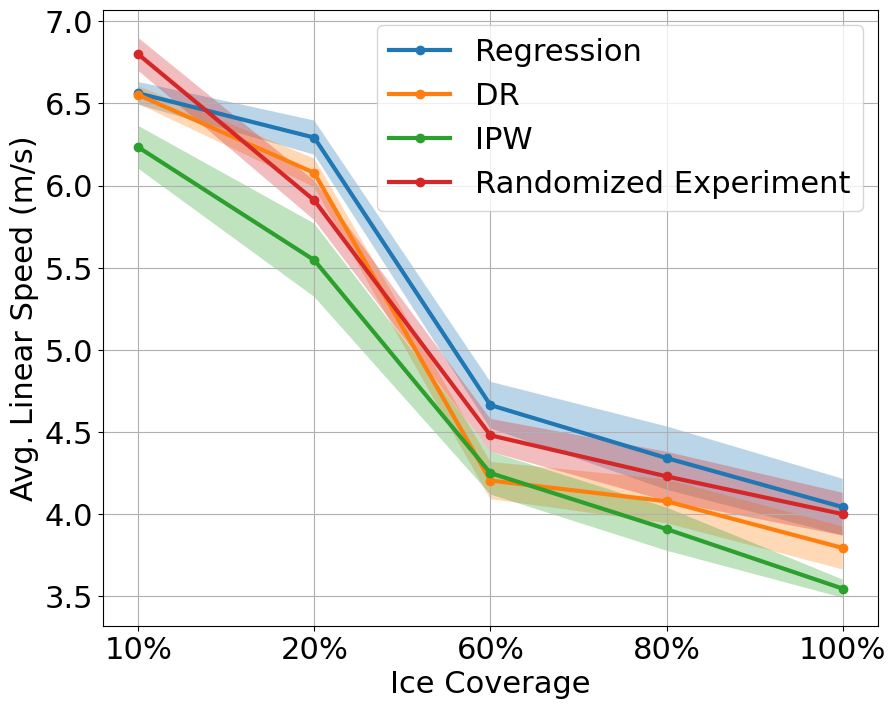}}
    \hspace{0.2in}
    \subfloat[]{\label{fig:angular-v}\includegraphics[width=0.25\linewidth]{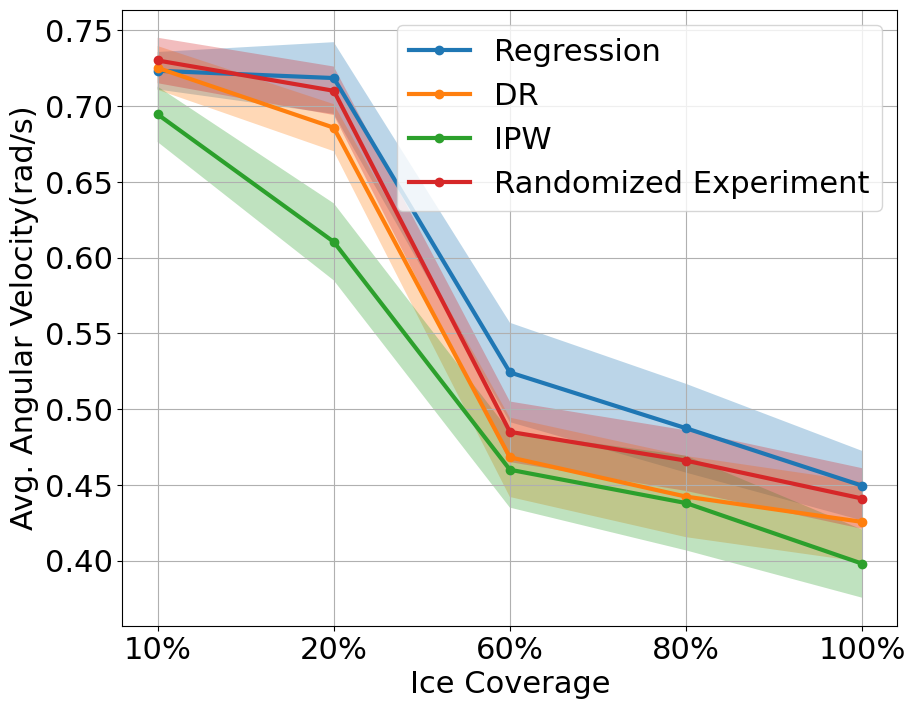}}\vspace{-10pt}
    \caption{\small Statistics for action executions of the three methods.
    In all figures, the x-axis represents the ice coverage percentage.
    (a) The frequency of taking aggressive actions. These actions are defined as a set $\{a | v \geq 6 \land |\omega| \geq \frac{\pi}{2}\}$, where $v$ and $\omega$ are linear and angular speeds, respectively.
    (b) Average linear speeds for five environments;  (c) Average angular speeds.
    The statistics averaged over $50$ trials. \vspace{-10pt}
    } 
    \label{fig:action-statistics}
\end{figure*}

Regression models are commonly employed in practice to learn transition functions in model-based RL~\cite{moerland2020model}. 
This approach works well when the model is not misspecified or the aforementioned observational data issues are not a concern. 
By a misspecified model we mean that a wrong model form is used or the model does not include key features. 
On the other hand, if the propensity score model is misspecified, the IPW approach may not produce unbiased results. It is natural to ask whether it is possible to combine the IPW estimator and regression estimator such that if one of them is incorrectly specified, the other correctly specified estimator can ensure the final unbiased result. 
We adopt the {\em doubly robust (DR)} mechanism \cite{robins1994estimation} to achieve this goal and to develop the estimation methods for our problem. 


Suppose the following (non-parametric) regression is fitted in the subset of sample $\mathcal{N}(u)$ for all state shifts associated with action $a$, $\Delta s_i = f_a(u_i) + \epsilon_i, \mbox{ } i\in\mathcal{N}(u),$
where $f(\cdot)$ is a vector non-parametric function and a noise term $\epsilon_i$ follows the zero-mean  normal distribution. Denote the fitted regression by $\hat{f}$ and the estimation at $u$ by $\hat{\mu}_a^{NR}(u)$, i.e., $\hat{\mu}_a^{NR}(u) := \hat{f}_a(u)$.
The following estimator combines IPW and regression estimators for the first moment 
\begin{align}\label{DR-first}
   \resizebox{0.8\hsize}{!}{${\displaystyle \hat{\mu}^{DR}_a(u) = \frac{1}{|\mathcal{N}(u)|}\sum_{i\in\mathcal{N}(u)}\left[\frac{\mathbb{I}_{a}(a_{i})\Delta s_{i}}{\hat{e}_{a}(u_i)}+ \left(1- \frac{\mathbb{I}_{a}(a_{i})}{\hat{e}_{a}(u_i)}\right)\hat{\mu}_a^{NR}(u_i)\right]}$}.
\end{align}
This estimator possesses the DR property: if the propensity score model is incorrect, the estimation is still unbiased as long as the regression model is correct. The vice versa is also true. 
Because $\mathbb{E}(\Delta s_i \Delta s_i^T) = \mathbb{E}(\epsilon\,\epsilon_i^T) + f_a(u_i)f_a^T(u_i)$, the regression estimator for the second moment of the transition function can be written as $\hat{\sigma}_a^{NR}(u) := \frac{1}{|\mathcal{N}(u)|}\sum_i( \hat{f}_a(u_i)\hat{f}_a^T(u_i) + \hat{\epsilon}_i\,\hat{\epsilon}_i^T )$ where $\hat{\epsilon}_i$ denotes the residuals. In practice, the product $\hat{f}\hat{f}^T$ is usually the dominant term and can be used as an approximate second-moment estimate. The DR estimator $\hat{\sigma}_a^{DR}(u)$ is defined similarly 
\resizebox{.68\hsize}{!}{${\displaystyle \hat{\sigma}^{DR}_a(u) = \frac{1}{|\mathcal{N}(u)|}\sum_{i\in\mathcal{N}(u)}\left[\frac{\mathbb{I}_{a}(a_{i})\Delta s_{i}\Delta s_{i}^T}{\hat{e}_{a}(u_i)}+ \left(1- \frac{\mathbb{I}_{a}(a_{i})}{\hat{e}_{a}(u_i)}\right)\hat{\sigma}_a^{NR}(u_i)\right]}$}.
Finally, to compute the optimal value function, we first apply the proposed causal inference-based methods in Sect.~\ref{sec:IPW} and ~\ref{sec:DR} to learn the first and second moments of the state shift at each supporting state. 
Second, we solve the diffusion approximated MDP via the kernel methods described in Sect.~\ref{DiffApproxMDP}. 

\vspace{-6pt}
\section{Simulated Experiments}\label{sec:simulated-exp}

\textbf{Proof-of-Concept Experiment:}
We first conduct experiments in a fully-controllable simulated environment to evaluate the advantage of our method without considering the full complexity of the real-world experiments.
The task of the robot is to drive as fast as possible on a cross-terrain elliptical track consisting of ice, concrete, and pebbles. 
In the simulation, the mobile robot dynamics is similar to the one used in the \textit{highway-env} environment~\cite{highway-env}, but this model is not known to the robot and needs to be learned. 
To generate the observational data, we use the behavior policy computed using an unbiased dataset (by randomizing the actions) and a safety control strategy to collect the observational data.
Although this data collection strategy ensures the safety of the vehicle, it also introduces a large bias.

\begin{figure}[t]
  \begin{center}
    \includegraphics[width=0.65\textwidth]{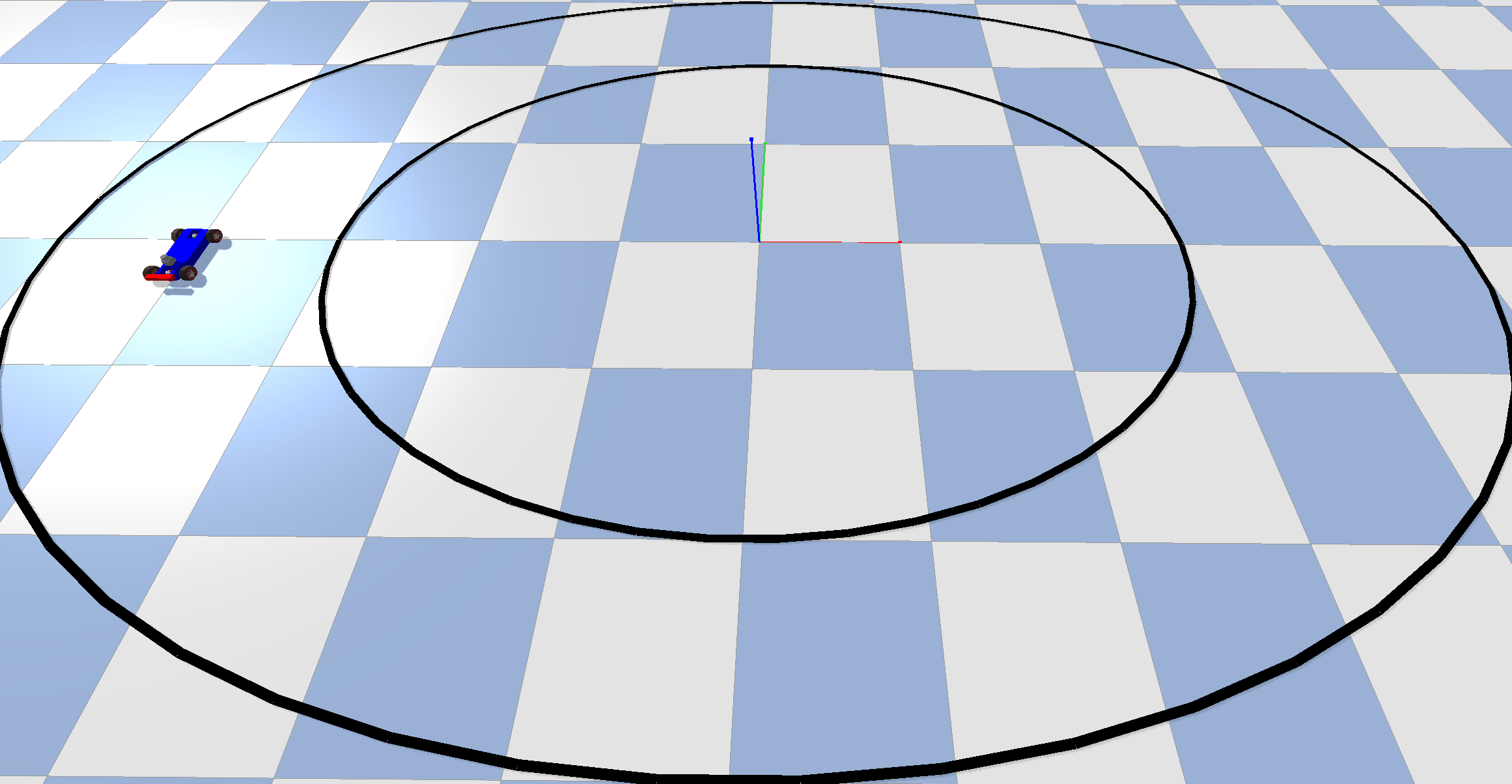} \vspace{-10pt}
  \end{center}
  \caption{\small Race car environment in the PyBullet simulator.\vspace{-10pt}} 
\end{figure}

\begin{table*}[t]
\caption{\small{Performance comparison for the PyBullet experiment. The scores are normalized according to $\frac{R^{\pi} - R^{rand}}{R^{*} - R^{rand}}$, where $R^{\pi}, R^{rand}, R^{*}$ are cumulative rewards for the policy computed using the learned model, randomized policy, and the policy computed by the ground-truth simulator, respectively. IL stands for imitation learning.}}
\resizebox{0.8\textwidth}{!}{%
\begin{tabular}{c|c|c|c|c|c|c|c|c}\toprule
   & Regression & IPW  & DR   & MOReL~\cite{kidambi2020morel} & IPW-MOReL & DR-MOReL & Q-Learning & IL\\ \midrule 
\rule{0pt}{5pt} $m=15kg$ & 0.64       & 0.72 & 0.75 & 0.71  & 0.81 & 0.82 & 
0.24 & 0.33  \\ 
\rule{0pt}{5pt} $m=20kg$ & 0.72       & 0.73 & 0.84 & 0.85  & 0.86  & 0.92 & 0.33 & 0.39   \\ 
\rule{0pt}{5pt} $m=25kg$ & 0.75       & 0.78 & 0.81 & 0.79  & 0.87      & 0.91 & 0.38 &  0.41  \\ 
\rule{0pt}{5pt} $m=30kg$ & 0.71       & 0.83 & 0.89 & 0.88  & 0.95      & 0.97 & 0.51 & 0.52    \\ \bottomrule
\hline
\end{tabular}
\label{tab:bullet-result}
}
\end{table*}

We compare the performance of the policy generated via the planner introduced in Sect.~\ref{DiffApproxMDP} using the model learned by our proposed estimators (inverse propensity score and doubly robust) with the standard regression method.
To make a fair comparison, we use the k-nearest neighbors (KNN) for the standard regression method, which is the same as the $\hat{\mu}_a^{NR}(u)$ in the doubly robust method in Eq.~\eqref{DR-first}.  
We run $50$ trials in each environment with 
$20000$ time steps and then average statistics. 

We first compare the cumulative undiscounted reward averaging over $50$ runs as shown in Fig.~\ref{fig:reward}. 
In addition to the standard regression approach, we also compare to the behavior policy that uses the unbiased dataset obtained by randomized action selection. 
We can observe that the regression method achieves slightly better performance when the ice coverage is below $20\%$.
However, as soon as the environment becomes more challenging, i.e., ice coverage greater than $60\%$, the causal inference-based methods start to excel, and are comparable with the model using the unbiased dataset!
This is because the biased data mostly exist on the ice surfaces due to the intervention during data collection.
As a result, when the ice coverage becomes larger, the regression method uses more biased data, and the performance degrades drastically. 
The comparison of the action execution statistics among the three methods shown in Fig.~\ref{fig:action-statistics} also reveals why the causal-based methods can achieve better performance. 
Fig.~\ref{fig:action-statistics}\subref{fig:aggressive-action} shows the average number of aggressive actions (with large linear and angular velocities simultaneously) taken by each method in the five environments. 
The general trend for all three methods shows that the robot motion becomes less aggressive when the environment is covered by more ice. 
In general, the causal-based methods take less aggressive actions than the regression-based model, and this is the desired behavior due to the safety concern. 
This phenomenon also explains why IPW and DR achieve more stable performances and can outperform regression in challenging environments. 

\vspace{-5pt}
\textbf{Experiments in Physics-Engine Simulator:}
\begin{table*}[t]
{\small
\centering
\caption{\small{Comparison between DR (doubly robust) and RG (regression) for the field experiments.}}
\label{tab:field-exp}
\begin{tabularx}{0.75\textwidth}{{c}p{0.3\textwidth}XXXXXX}\toprule
     & \multicolumn{2}{|c}{Success Rate (\%)} & \multicolumn{2}{|c}{Traversal Time (s)} & \multicolumn{2}{|c}{Average $\theta_y$ (degree)}  \\ \midrule
\rule{0pt}{3pt} Waypoints   & \multicolumn{1}{|c}{DR} & \multicolumn{1}{c}{RG} & \multicolumn{1}{|c}{DR} & \multicolumn{1}{c}{RG} & \multicolumn{1}{|c}{DR} & \multicolumn{1}{c}{RG} \\ \midrule
\rule{0pt}{3pt}
A & \multicolumn{1}{|c}{$\bf{0.8}$} & \multicolumn{1}{c}{$0.5$} & \multicolumn{1}{|c}{${23.73 \pm 1.28}$} & \multicolumn{1}{c}{$\bf{21.62 \pm 2.11}$} & \multicolumn{1}{|c}{$\bf{1.91 \pm 0.67}$} &  \multicolumn{1}{c}{$2.93 \pm 0.81$}\\ 
\rule{0pt}{3pt}
B & \multicolumn{1}{|c}{$\bf{0.6}$} & \multicolumn{1}{c}{$0.3$} & \multicolumn{1}{|c}{$\bf{21.76 \pm 3.88}$} & \multicolumn{1}{c}{$26.27 \pm 4.6$} &    \multicolumn{1}{|c}{$\bf{3.48 \pm 1.2}$} & \multicolumn{1}{c}{${4.79 \pm 0.77}$}  \\ 
\rule{0pt}{3pt}
C & \multicolumn{1}{|c}{$\bf{0.9}$} & \multicolumn{1}{c}{${0.8}$} & \multicolumn{1}{|c}{$34.66 \pm 1.7$} & \multicolumn{1}{c}{$\bf{31.66 \pm 2.21}$}     &     \multicolumn{1}{|c}{$\bf{2.41 \pm 0.46}$} & \multicolumn{1}{c}{$2.42 \pm 1.02$}     \\ \bottomrule
\end{tabularx}
}
\vspace{-5pt}
\end{table*}
We further evaluate our method in PyBullet~\cite{coumans2021} which is a high-fidelity physics simulator.
The observational data collection process is similar to the previous section, but here we use the vehicle's body mass as a feature.
In addition to comparing with the standard regression method, we also include a comparison with an offline model-based method, MOReL~\cite{moerland2020model}, and two methods that combine causal-based de-biasing with MOReL, i.e.,  IPW-MORel and DR-MOReL. 
This comparison is to show that our de-biasing method can be used to improve offline model-based RL. 
MOReL categorizes the state-action pairs as unknown if they are not frequently appeared in the dataset, and adds a large penalty to them. 
The original MOReL uses the disagreement of deep ensembles to decide which state-action pairs to be viewed as unknown.
In our experiment, we use a similar approach: we detect the unknown state-feature-action tuples by using the number of examples within an $\epsilon$ ball of the current queried point. 
We also include two model-free methods, Q-learning~\cite{sutton2018reinforcement} and imitation learning (IL)~\cite{hussein2017imitation}.
We use the same observational dataset described above to train all the methods.
Since we do not allow the robot to interact online with the environment, the Q-learning method is trained only on the static dataset collected by the behavioral policy.
Because PyBullet is a deterministic simulator, we add a small Gaussian noise to the actions before execution to simulate stochastic effects.
The normalized cumulative rewards for different methods are shown in Table~\ref{tab:bullet-result}.
We can see all model-based methods outperform the model-free ones.
We conjecture that learning the first and second moments of the motion model is easier than learning a Q function or a policy from only the offline data due to the distributional shift~\cite{de2019causal, kumar2020conservative}.
We can also observe the causal inference-based methods perform better than the standard regression estimator. 
Specifically, the DR estimator generates the best results across all different vehicle masses in the uncombined methods.
By comparing the combined method (IPW-MOReL and DR-MOReL), we observe that the two methods can leverage the strengths of each other and outperform all other methods.
We conclude the causal inference-based method provides a better estimate of the motion model from offline data, and the planner leverages this estimation to compute a better policy.

\vspace{-5pt}
\section{Hardware Experiments}

\begin{figure}[t] \vspace{-10pt}
    \centering
    {\label{fig:}\includegraphics[width=0.45\linewidth, height=1.1in]{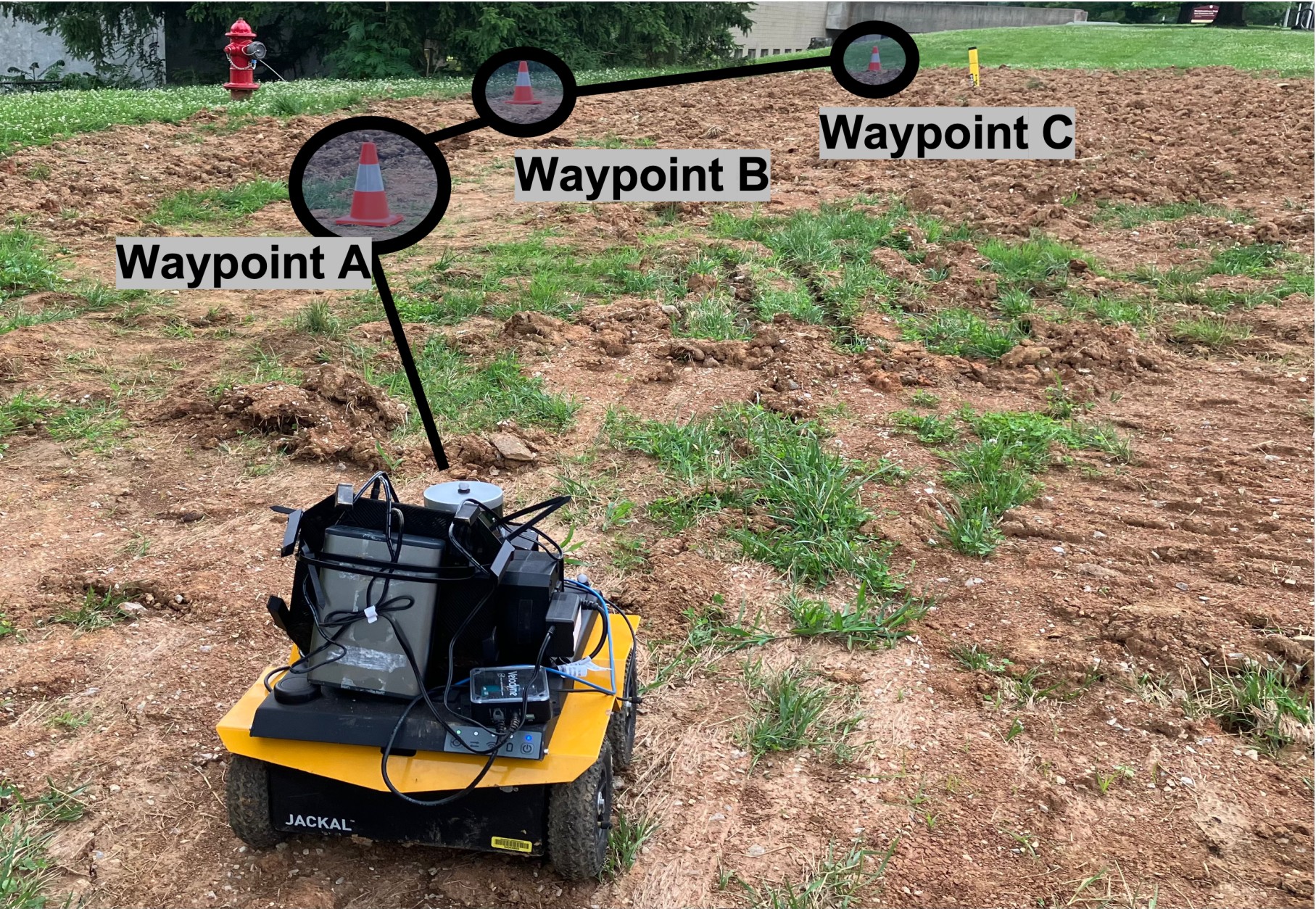}}
    \hspace{10pt}
    {\label{fig:elevation}\includegraphics[width=0.45\linewidth, height=1.1in]{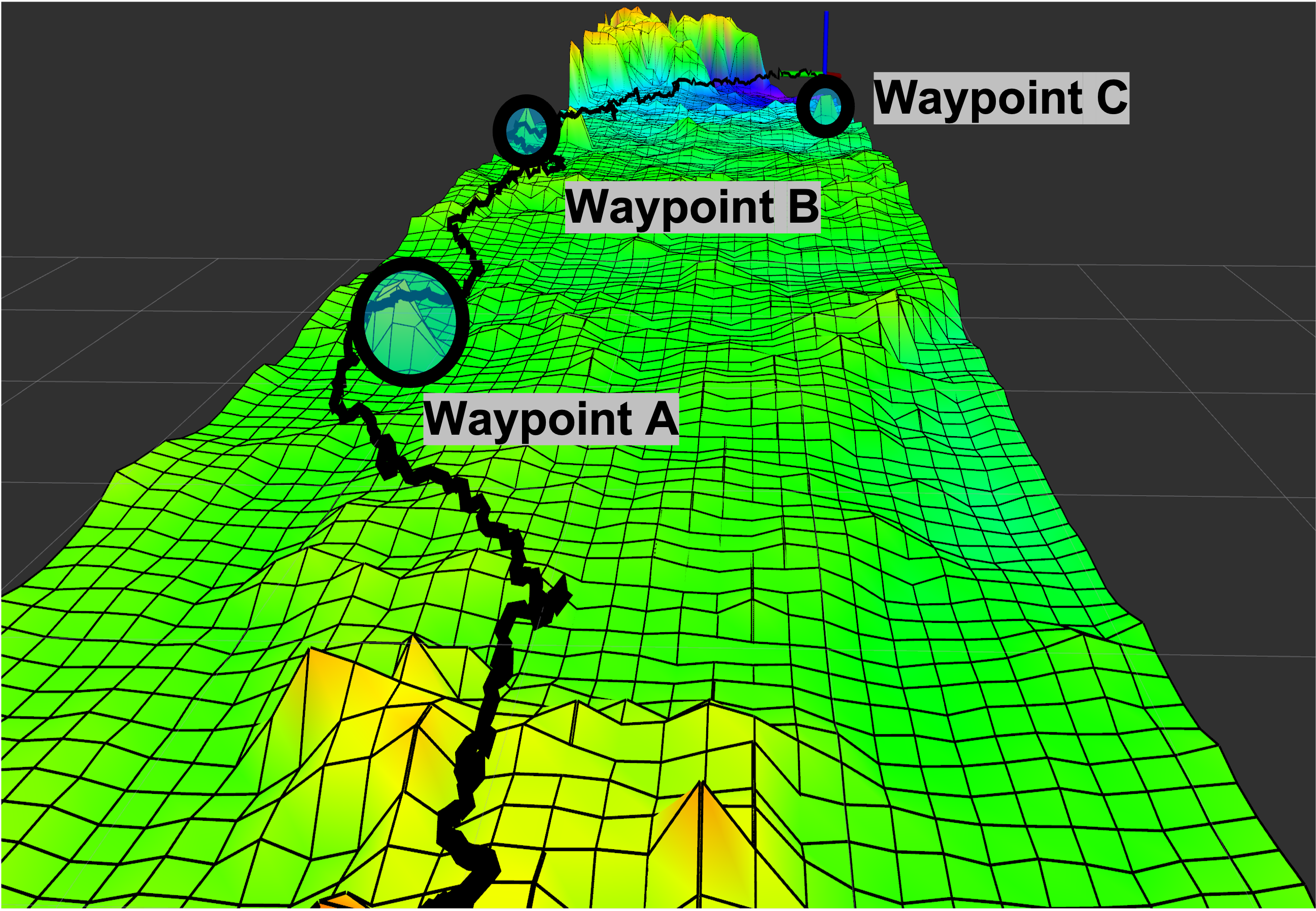}}
    \caption{\small Field experiment using a ClearPath Jackal. 
    Left: The environment is an uneven dirt field. 
    Right: Visualization of the elevation map constructed using LiDAR points shown in the 2.5D color map. 
    The black trajectory is a successful run of our method navigating through three waypoints. \vspace{-12pt}
    } 
    \label{fig:real-world-exp}
\end{figure}
We further conducted hardware experiments using a Jackal ground vehicle navigating in an off-road environment.
The environment and its elevation map, which is used as the feature for estimating the motion moments, are shown in Fig.~\ref{fig:real-world-exp}.
This environment contains dirt fields with uneven surfaces. 
The robot's task is to arrive at three pre-defined  waypoints sequentially ($A\rightarrow B \rightarrow C$).
We compare the DR version of the causal-based method with a regression model which does not have causal-based bias correction. 
We perform $10$ experiments for each method and use the following three commonly-used metrics for robotic navigation tasks to evaluate the performance~\cite{siva2021enhancing}. 
\begin{itemize}
    \item Success rate: the success rate indicates the ratio between successful trials and the total trials. 
    A trial is successful if the robot can navigate from the starting position to the ending position within $0.5m$.
    If the robot cannot move, e.g., due to flipping, the run is viewed as a failure. 
    \item Travel time: these metric averages travel time (in seconds) over the successful trials. 
    It measures the efficiency of the robot in completing the navigation task. 
    \item Averaged pitch angle: 
    it measures the average pitch angle of the robot over one successful experiment.   
    It is crucial to keep the pitch angle low 
    to avoid flipping. \vspace*{-5pt}
\end{itemize}

We summarize the results of the above metrics of the DR method and the regression method in Table~\ref{tab:field-exp}.
Each row shows the performance of navigating to the corresponding waypoint from the previous one. 
The DR method has a consistently better success rate and smaller average pitch angle than regression for all the waypoints.
It is worth noting that although the regression-based method outperforms DR in terms of the traversal time when navigating to waypoints A and C in successful runs, the low success rate and large $\theta_{y}$ indicate the unreliable and aggressive performance of the regression-based method due to the bias in the dataset. 
Additionally, compared to the standard regression method, we observe that the reason for a slightly longer traversal time for DR is that it tried to avoid many inclined terrains to keep safe, which caused it to deviate from the straight line path between two waypoints.
This observation is consistent with the simulated experiments' results -- the causal-based methods can provide better estimates by properly handling the bias in the dataset, resulting in safer and more reliable planning performance.

\vspace{-7pt}
\section{Conclusion}
We present a principled framework by synthesizing causal inference with a diffusion-approximated MDP for 
robot decision-making with unknown motion model parameters.
It enables the robot to compute a correct policy by learning only the first two moments of the stochastic transition model from biased observational data, which is data-efficient (only learning important statistics) and behaviorally safe (no random robot explorations).
We conducted extensive experiments in both simulation and real world, and the results show our method can
learn the parameters of the motion model correctly and efficiently in challenging environments, and is evidently superior to the conventional regression-based framework in terms of de-biasing and utilizing offline data. 

{\small
\bibliographystyle{IEEEtran}
\bibliography{ref}
}
\clearpage
\end{document}